\documentclass[lettersize,journal]{IEEEtran}
\usepackage{amsmath,amsfonts}
\usepackage{algorithmic}
\usepackage{algorithm}
\usepackage{array}
\usepackage[caption=false,font=normalsize,labelfont=sf,textfont=sf]{subfig}
\usepackage{textcomp}
\usepackage{stfloats}
\usepackage{url}
\usepackage{verbatim}
\usepackage{graphicx}
\usepackage{cite}
\usepackage{orcidlink}

\usepackage{booktabs}
\usepackage{multirow}
\usepackage[table]{xcolor}
\definecolor{tableorange}{HTML}{FFCC33} 
\usepackage{float}
\usepackage{bm}
\usepackage{cuted}
\usepackage{color,soul}
\begin{document}

\bstctlcite{IEEEexample:BSTcontrol}


\title{Vision-Reasoning-Guided Occlusion Removal from Light Fields}

\author{
Mohamed Youssef \,\orcidlink{0009-0005-2031-8432},
and Oliver Bimber \,\orcidlink{0000-0001-9009-7827}%
\thanks{This research was funded by Austrian Science Fund (FWF) Grant 32185-NBL “Wide Synthetic Aperture Sampling”}
\thanks{Mohamed Youssef and Oliver Bimber are with the Department of Computer Science, Johannes Kepler University, Altenbergerstr. 65, Linz, 4040, Austria. (e-mail: mohamed.youssef@jku.at; oliver.bimber@jku.at)}
\thanks{Corresponding author: Oliver Bimber (e-mail: oliver.bimber@jku.at).}
}

        





\maketitle

\begin{abstract}

Occlusion-robust scene recovery remains a major challenge in computational imaging, particularly where dense vegetation severely limits visibility. We propose a vision-reasoning-guided light field occlusion removal framework combining light field integration (LFI) with vision-language model (VLM) semantic reasoning. Multi-view observations are first integrated via LFI to suppress foreground occlusions, producing an initial visibility-enhanced representation, a VLM then acts as a conditional semantic prior to restore degraded structures and fine details. A multi-sample fusion strategy aggregates multiple generated hypotheses to improve consistency and reduce hallucination. Experimental results on synthetic and real-world datasets show state-of-the-art performance, achieving the highest average SSIM across four synthetic benchmark scenes (4-Syn) and strong generalization across structured and unstructured acquisition settings, with applicability to search-and-rescue and exploratory robotic navigation.

\end{abstract}


\section{Introduction}

\IEEEPARstart{C}{\MakeLowercase{omputational}} visual perception systems have achieved remarkable progress in object detection \cite{od_1, od_2, od_3}, tracking \cite{ot_1, ot_2, ot_3}, and segmentation \cite{seg_1}. Nevertheless, their performance remains severely limited in natural environments, where dense vegetation frequently occludes objects of interest, degrading the visual evidence available for scene interpretation. This challenge extends beyond military applications to civilian domains including search and rescue \cite{aos_1, aos_2}, autonomous robot navigation \cite{aos_3}, wildlife monitoring \cite{aos_4}, wildfire assessment \cite{aos_5}, and forest ecology analysis \cite{aos_14}, where recovering the underlying scene from heavily occluded observations is fundamentally ill-posed.

\begin{figure}
\centering
\includegraphics[width=3.5in]{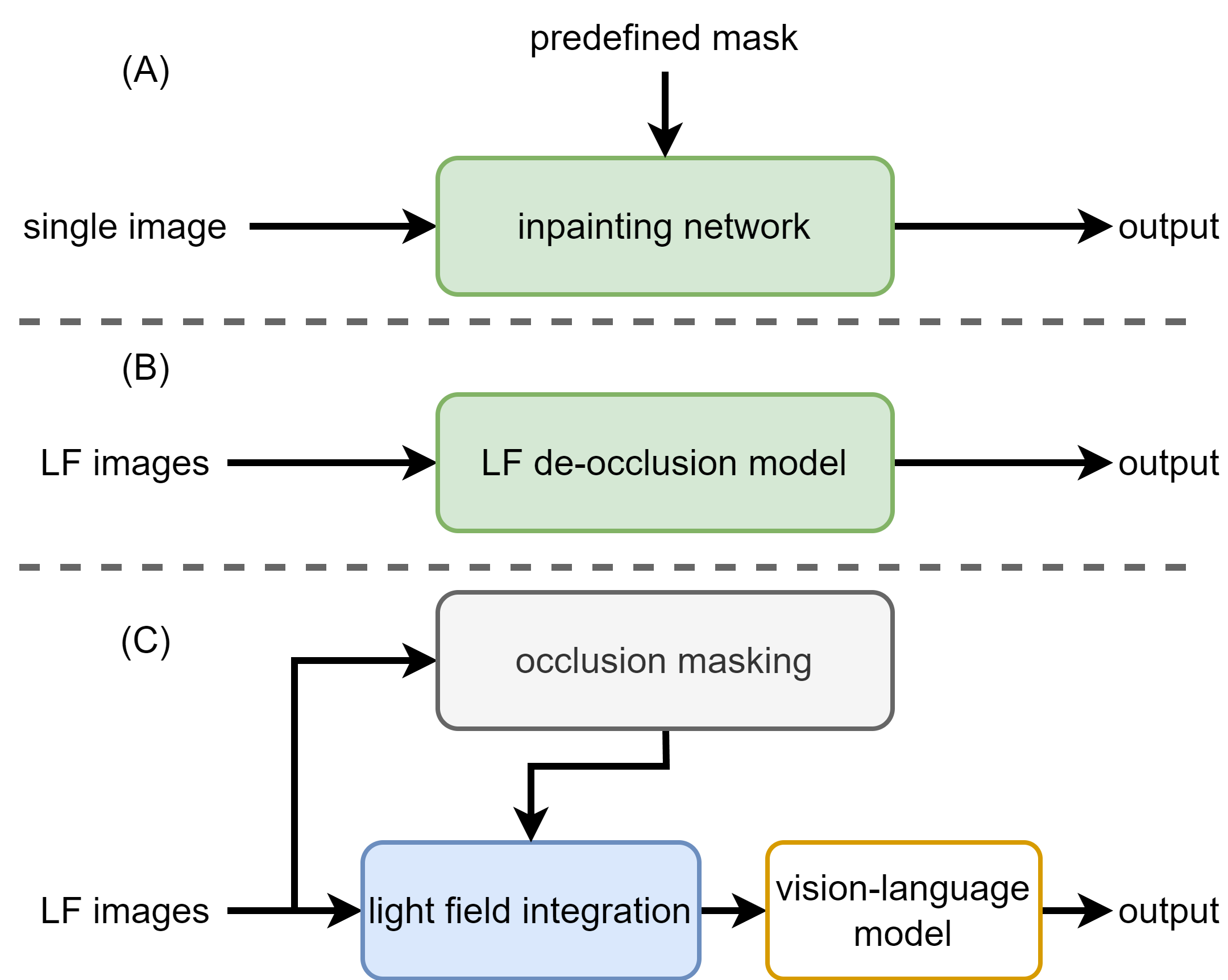}
\caption{Framework comparison between existing single-image occlusion removal methods (A), existing light field (LF) occlusion removal methods (B), and the proposed framework (C).}
\label{fig_1}
\end{figure}

Single-image occlusion removal, commonly formulated as image inpainting, attempts to recover missing content from incomplete observations. However, under severe vegetation occlusion, hidden regions are often never directly observed, forcing reconstruction to rely primarily on learned semantic priors that may hallucinate structures inconsistent with the true scene geometry. Many inpainting methods also require predefined occlusion masks, as illustrated in Fig.~\ref{fig_1}(A), which are difficult to obtain reliably in complex outdoor environments.

Light field (LF) imaging alleviates this limitation by exploiting multi-view observations: sampling both spatial and angular information allows background regions occluded in some viewpoints to become visible in others, enabling visibility enhancement through light field integration (LFI) and digital refocusing.

Existing LF occlusion-removal approaches fall broadly into signal-processing and deep learning methods. Signal-processing approaches exploit geometric consistency and synthetic apertures for multi-view integration but often suffer from residual blur and incomplete occluder suppression. Learning-based methods, illustrated in Fig.~\ref{fig_1}(B), use deep neural networks to separate foreground occluders from the underlying scene via parallax and visibility cues, however, their performance degrades under severe occlusions, with limited generalization to sparse, noisy, or unstructured viewpoint configurations.

Recent vision-language models (VLMs) have shown strong semantic reasoning and image generation capabilities \cite{image_gen_1, image_gen_2}, enabling plausible inference of scene structure from incomplete observations--particularly attractive for LF reconstruction, where severe occlusions leave parts of the scene weakly constrained. However, as generative models, VLMs may produce hallucinated content unsupported by the captured measurements, limiting their direct use in physically grounded reconstruction.

Despite these advances, existing methods do not jointly exploit the complementary strengths of LFI and semantic vision reasoning for occlusion-robust scene reconstruction, as illustrated in Fig.~\ref{fig_1}(C), leaving perceptually coherent reconstruction under sparse, partial observations an open challenge.
To address this, we propose a vision-reasoning-enhanced light field occlusion removal framework, as illustrated in Fig.~\ref{pipeline_overview}. The framework first applies LFI to suppress foreground occluders through geometrically consistent multi-view aggregation, producing an initial visibility-enhanced reconstruction. A vision-language model then refines this reconstruction by restoring structural and semantic details in degraded regions. Finally, a multi-sample fusion strategy averages multiple independently generated reconstructions to suppress hallucination artifacts and improve measurement consistency. By combining physical imaging constraints with semantic vision reasoning, the proposed framework achieves more faithful and robust reconstruction under severe occlusion.

\section{Related works}

In this section, we first review single-image inpainting methods. Next, we discuss traditional LF occlusion removal approaches. Finally, we present an overview of state-of-the-art deep learning-based methods for LF occlusion removal.

\subsection{Single-Image Inpainting}

Single-image inpainting methods aim to predict pixel values within masked regions to remove foreground occlusions \cite{single_image_18, single_image_19, single_image_20, single_image_21, single_image_22}. These methods typically reconstruct missing content by leveraging information from neighboring pixels or global image priors. However, since single 2D images lack explicit depth information, occlusion masks must be predefined, and the accuracy of occlusion identification depends heavily on the volume and diversity of training data. In contrast, LF imaging captures multiple views of a scene, enabling automatic identification of foreground occlusions through analysis of the underlying scene geometry: occluding objects, being closer to the camera than the background, exhibit larger disparities across views.

\begin{figure*}[!t]
\centering
\includegraphics[width=7.5in]{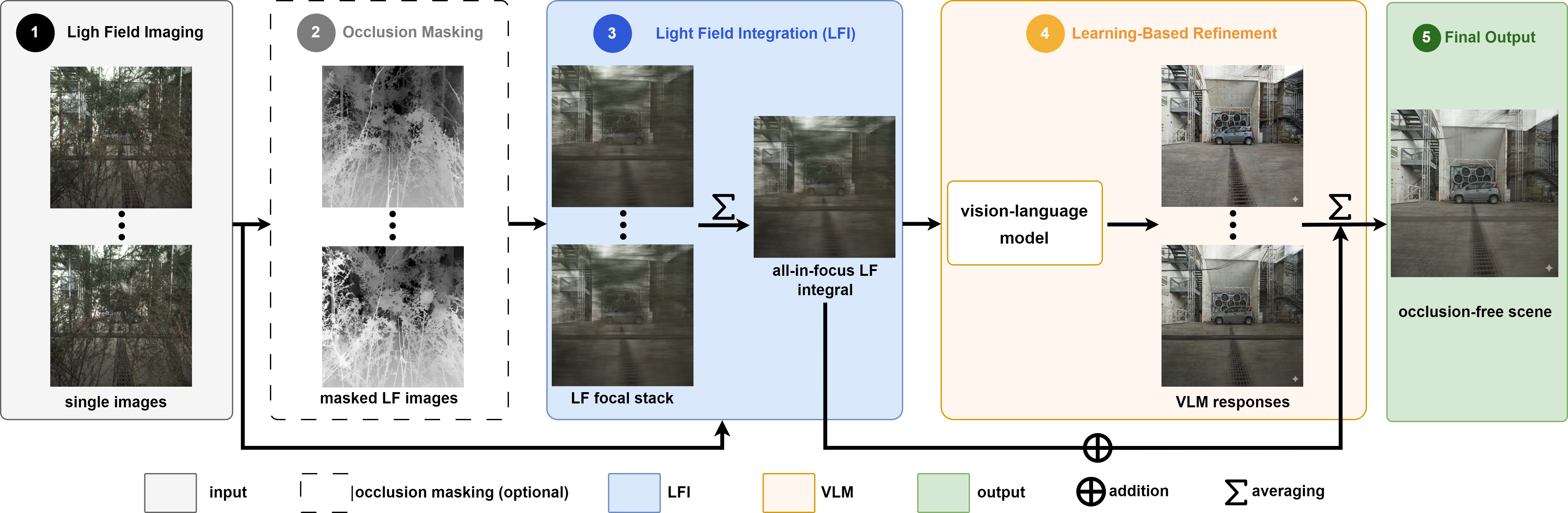}
\caption{Overview of the proposed hybrid occlusion-free reconstruction framework. The framework consists of four main stages to recover occlusion-free scene representations in highly vegetated environments. First, multi-view images are acquired from diverse viewpoints. An optional occlusion masking preprocessing step then suppresses dominant foreground occluders to produce an initial occlusion-reduced representation. Next, LFI aggregates multi-view observations in a geometrically consistent manner to mitigate residual occlusions and reconstruct underlying scene content. Finally, a VLM refines high-frequency details, enhances perceptual quality, and produce an occlusion-free scene representation.}
\label{pipeline_overview}
\end{figure*}

\subsection{Traditional LF Occlusion Removal}

Occlusion removal in LF imaging is an active research topic that has been investigated for several decades \cite{TR_1, TR_2, TR_3, TR_4}. A LF consists of multiple perspective images that capture both the angular domain and the spatial domain, resulting in a four-dimensional (4D) representation of the scene. Occlusion removal in LF images relies on complementary information from multiple views to recover scene content that is hidden behind foreground occlusions.

For example, \cite{TR_1, TR_2, TR_3} proposed refocusing LF images to suppress foreground occlusions by blurring them while keeping the background in focus. In this approach, the refocused image is generated by averaging images from different views, producing blurred objects away from the focal plane and sharp objects on the focal plane. The challenge of removing the defocus blur in synthetic aperture images to obtain an all-in-focus image has been further addressed in \cite{TR_4}.

\subsection{Deep Learning LF Occlusion Removal}
Deep learning has been widely applied to LF tasks, including super-resolution \cite{DL_1} and depth estimation \cite{DL_2}. DeOccNet \cite{DL_3} was the first deep learning model for LF occlusion removal, stacking all sub-aperture images to exploit occluded-object information and encoding them via an encoder–decoder architecture. However, stacking along a single dimension neglects spatial relationships between sub-aperture images, limiting reconstruction quality.

To address this, Mask4D \cite{DL_4} introduced a 4D convolution layer preserving spatial connections between sub-aperture images. Similarly, LFORNet \cite{DL_5} proposed an end-to-end framework using multi-angle view stacks, decomposing the task into sub-networks for foreground occlusion localization and background recovery. In \cite{DL_6}, a three-module architecture explicitly estimates occlusion masks from LF features to inpaint foreground occlusions. MANet \cite{DL_7} further introduced joint training of an occlusion mask predictor and remover, restoring backgrounds via gated spatial–angular feature aggregation with a texture–semantic attention mechanism.

In addition, \cite{DL_8} proposed a progressive multi-plane image (MPI) construction method that builds MPI layers sequentially from near to far, leveraging occlusion priors from nearer layers to recover hidden background. ELFNet \cite{DL_9} combines convolutional layers with a Swin Transformer \cite{DL_10} to exploit both local and global receptive fields.

Despite this progress, several limitations remain. Many state-of-the-art methods rely on convolution-based architectures, which primarily capture local features and struggle to model long-range dependencies across the LF's spatial and angular dimensions. While recent approaches incorporate transformers for global feature capture, their feature extraction modules generally do not explicitly distinguish occlusion pixels from background pixels. Furthermore, existing methods assume a fixed angular input size, limiting flexibility for real-world applications where the number of available views may vary.

\section{Proposed method}
In this section, we introduce the proposed approach: a hybrid reconstruction framework that integrates LFI with VLMs to recover occlusion-free scene representations under severe visibility constraints, as illustrated in Fig.~\ref{pipeline_overview}.

The pipeline consists of four main stages. First, multi-view images are acquired from environments heavily occluded by vegetation, using platforms such as drones, mobile robotic systems, or LF cameras, yielding either structured or unstructured LF data. Second, an optional occlusion masking stage suppresses foreground occluders, producing an initial occlusion-reduced representation. Third, LFI performs geometrically consistent aggregation of multi-view observations to suppress foreground occlusions and recover underlying scene content, however, the resulting intermediate reconstruction may still exhibit defocus blur, noise, and loss of fine detail due to limited visibility and incomplete observations.

To address these limitations, we integrate a VLM as a conditional refinement module that introduces strong semantic priors for recovering missing structures and enhancing degraded regions. Operating on the LFI reconstruction, the VLM refines high-frequency details and improves perceptual quality using semantic and structural representations learned from large-scale visual data. Rather than a standalone post-processing step, the VLM serves as a semantically guided prior that complements the reconstruction process, enabling visually coherent and semantically plausible scene content even where the underlying signal is severely degraded, partially missing, or fundamentally limited by occlusion.

\subsection{LF Imaging}
In optical imaging, LF acquisition falls broadly into structured and unstructured viewpoint sampling. Structured sampling captures viewpoints on a predefined regular grid using camera arrays  or lenslet-based plenoptic cameras, simplifying geometric calibration, digital refocusing, and depth estimation. However, it is constrained by the physical aperture of the array, the spatial–angular resolution trade-off inherent to plenoptic cameras, and precise hardware calibration and synchronization requirements.

Unstructured sampling instead acquires images from arbitrary viewpoints, as in mobile robotics and drone-based imaging. Since camera positions are not predefined, poses must be estimated via structure-from-motion (SfM) \cite{colmap}, SLAM \cite{slam}, or learning-based approaches such as VGGT \cite{vggt} or depth anything 3 \cite{da3}. Integrating images along a moving trajectory enables much larger synthetic apertures, suiting applications such as search and rescue, wildlife monitoring, and aerial ecological surveys \cite{aos_1,aos_2,aos_4,aos_5}. However, the quality of the resulting reconstruction depends critically on the accuracy of the underlying pose estimates.

Unlike existing state-of-the-art methods, typically restricted to structured grid sampling and synthetic sparse occlusions, the proposed approach supports both structured and unstructured acquisition, substantially improving applicability and generalization to real-world imaging scenarios.

\subsection{Occlusion Masking}
Occlusion masking constitutes an optional preprocessing stage within the proposed pipeline, intended to provide an initial suppression of foreground occluders when a suitable masking approach is available. Both classical and learning-based vegetation segmentation methods can be employed for this purpose. Conventional vegetation indices \cite{v_index}, may be utilized for appearance-based foreground suppression, while modern learning-based approaches, including depth anything 3 \cite{da3} and pixel-perfect depth \cite{ppd}, can provide more robust occlusion masking segmentation under complex environmental conditions. By suppressing foreground vegetation prior to reconstruction, the occlusion masking stage improves the quality and consistency of the subsequent LFI process, thereby facilitating more reliable recovery of the underlying scene content.

\subsection{Light Field Integration}
Light Field Integration (LFI) reconstructs images corresponding to a large synthetic aperture by integrating images acquired from multiple viewpoints. These images sample the four-dimensional light field (LF), parameterized as

\begin{equation}
\label{lf_equation}
L(x,y,u,v),
\end{equation}

where (x,y) denote the spatial image coordinates and (u,v) represent the angular coordinates corresponding to the viewpoint positions.

A scene point observed from different viewpoints exhibits a disparity-dependent displacement across the angular dimensions (u,v). This geometric relationship forms the basis of photogrammetric reconstruction, where feature correspondences are used to estimate scene depth. Under severe occlusion, however, feature visibility becomes discontinuous across viewpoints, resulting in missing correspondences and degraded depth estimation.

Given the estimated disparity, each captured view is shifted to align the projections of a target depth plane before integration. Averaging the aligned views produces a synthetic aperture image in which structures at the selected depth are sharply focused, while foreground occluders and out-of-focus regions are attenuated. Repeating this process over multiple focal depths and combining the resulting focal stacks yields an all-in-focus integral image that preserves structures across the entire scene while reducing defocus blur.

This LF formulation enables computational imaging tasks including digital refocusing, depth estimation~\cite{depth_2}, and occlusion suppression through Airborne Optical Sectioning (AOS)~\cite{aos_1, aos_2, aos_3, aos_4, aos_5, aos_6, aos_7, aos_8, aos_9, aos_10, aos_11, aos_12, aos_13, aos_14, aos_15, aos_16, aos_17, aos_18, aos_19, aos_20, aos_21, aos_22}.

\subsection{Learning-Based Refinement}

VLMs have demonstrated strong capabilities in understanding and reasoning about complex scene structures, enabling them to infer high-level semantic and geometric relationships from degraded visual inputs. However, their generative nature lacks guarantees of physical consistency and may introduce hallucinated content that is not supported by the underlying observations. To address this limitation, we employ VLMs, namely Gemini 3.1 and Qwen-Image \cite{qwen} as a conditional semantic prior to refine the intermediate LF integral image. Specifically, after foreground occlusions are suppressed via LFI, the VLM is used to recover high-frequency details and correct defocus artifacts, leveraging its learned semantic knowledge while remaining anchored to the input image. 

\begin{figure*}[t]
\centering
\includegraphics[width=\textwidth]{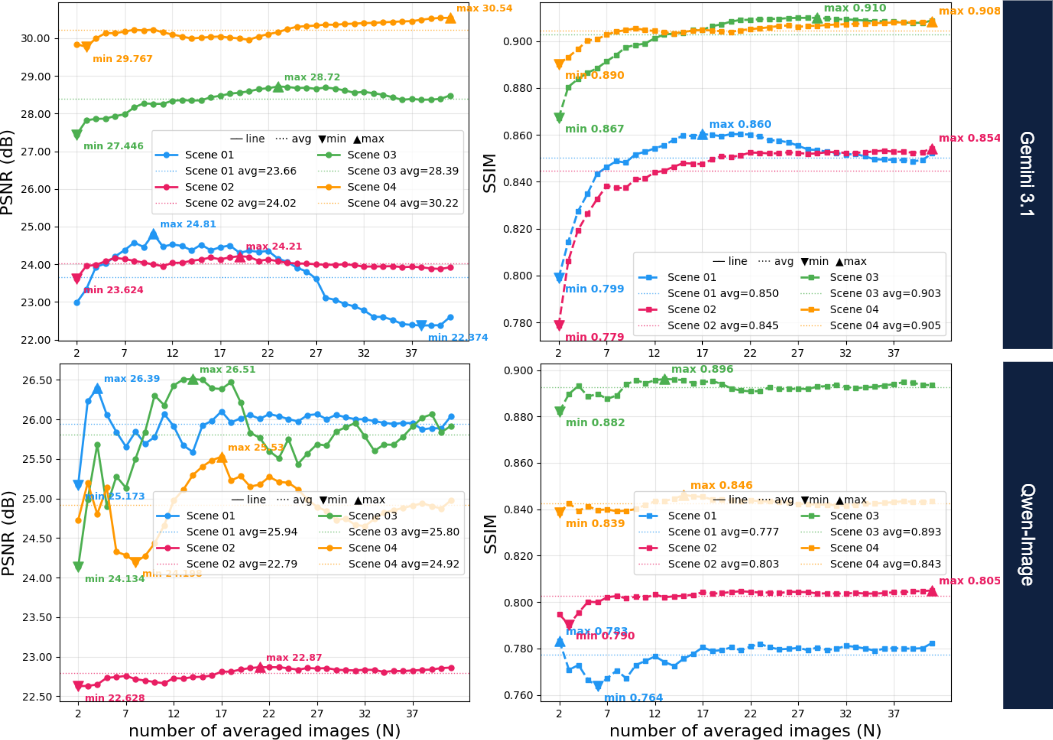}
\caption{Effect of the number of averaged generated images (N) on reconstruction quality for 4-Syn benchmark scenes using Gemini 3.1 (top) and Qwen-Image (bottom). PSNR (left) and SSIM (right) are shown as a function of N, dotted lines indicate scene-wise averages, while markers denote minimum and maximum values. SSIM generally improves and saturates around N = 20, whereas PSNR exhibits greater variability due to pixel-wise differences introduced during image generation.}
\label{vlm_stats}
\end{figure*}

To suppress hallucination artifacts that may arise from the VLM, we generate \(N\) independent responses and compute their pixel-wise average together with the all-in-focus LF integral to obtain the final reconstruction. The choice of \(N\) is empirically motivated based on experiments conducted using both state-of-the-art closed-source and open-source VLMs (cf. fig. \ref{vlm_stats}). 



Our experiment evaluates the relationship between the number of averaged generated images and the corresponding structural similarity index (SSIM) and Peak signal-to-noise ratio (PSNR) on the 4 synthetic benchmark scenes (4-Syn) proposed in \cite{DL_3}. The results demonstrate that the reconstruction quality progressively improves as additional generated samples are averaged. In particular, the SSIM exhibits a clear increasing trend before reaching saturation at approximately \(N\) = 20 averaged samples, beyond which no significant improvement is observed. This suggests an effective trade-off between reconstruction quality and computational cost while substantially reducing inconsistent hallucinated structures.

In contrast, the PSNR measurements exhibit a noisier trend across different numbers of averaged samples. This behavior can be attributed to pixel-level variations and local intensity alterations introduced by the VLM during the generation process. Since VLM-based reconstruction may introduce semantically plausible modifications and residual hallucinated content that are not strictly aligned at the pixel level, PSNR becomes more sensitive to such local deviations, even when the perceptual and structural quality of the reconstruction improves.

The VLM-guided reconstruction is formulated as:
\begin{equation}
\label{vllm_equation}
I_{\mathrm{VLM}} = \mathcal{G}\!\left(I_{\mathrm{LF}}, P\right)
\end{equation}
where $\mathcal{G}$ denotes the VLM inference function, $I_{\mathrm{LF}}$ is the integral image obtained from LFI, and $P$ represents a conditioning prompt that encodes the desired reconstruction constraints.

The prompt $P$ plays a critical role in guiding the reconstruction process. For the 4-Syn benchmark scenes, we employ the following prompt:
\textit{“Please remove the repetitive pattern noise and blur from the image. The processed image should remain aligned with the input image”}.

\section{results}

This section evaluates the quantitative and qualitative generalization capability of the proposed framework on synthetic and real-world datasets. On the synthetic dataset, our approach outperforms existing state-of-the-art LF occlusion removal methods in SSIM. On real-world scenes, where ground truth is unavailable, we rely on qualitative evaluation, which demonstrates robustness across diverse scenarios and highlights practical applicability to search-and-rescue and ecological monitoring.

\subsection{Quantitative Results}
Our approach delivers strong performance on LF 4-Syn benchmark scenes. This is demonstrated by the quantitative results in Table 1, where the PSNR and SSIM metrics highlight its effectiveness in occlusion removal.

In particular, our method achieves the highest average SSIM (0.883) among all compared methods, outperforming existing state-of-the-art approaches. This indicates that our reconstructed images preserve structural information and visual similarity more effectively. However, in terms of PSNR, which measures pixel-wise differences, our method achieves average PSNR (26.90), while the highest average PSNR (29.75) is obtained by Mask4D \cite{DL_4}.


\begin{table}
\caption{Quantitative comparison across Scenes 1--4. Bold values indicate the best performance, while underlined values represent the second-best performance.\label{tab:table1}}

\centering
\scriptsize
\setlength{\tabcolsep}{0pt} 
\begin{tabular*}{\columnwidth}{@{\extracolsep{\fill}}llccccc}
\toprule
\textbf{Metric} & \textbf{Method} & \textbf{scene 1} & \textbf{scene 2} & \textbf{scene 3} & \textbf{scene 4} & \textbf{Avg.} \\ 
\midrule
\multirow{15}{*}{{PSNR(dB) $\uparrow$}} 
 & All-In-Focus \cite{single_image_23} & 19.78 & 18.10 & 19.60 & 21.57 & 19.16 \\
 & RFR \cite{single_image_20} & 17.21 & 16.73 & 23.38 & 23.80 & 20.28 \\
 & Coordfill \cite{single_image_22} & 19.63 & 16.86 & 26.78 & 27.12 & 22.60 \\
 & GIP \cite{single_image_5} & 17.05 & 17.54 & 19.78 & 26.24 & 20.15 \\
 & PUT \cite{single_image_10} & 21.19 & 20.45 & \underline{31.24} & 24.45 & 24.33 \\
 & DeOccNet \cite{DL_3} & 23.10 & 18.92 & 23.23 & 26.14 & 22.85 \\
 & Mask4D \cite{DL_4} & \textbf{30.09} & \textbf{27.50} & 30.47 & \underline{30.93} & \textbf{29.75} \\
 & Zhang et al. \cite{DL_5} & 23.14 & 20.19 & 26.33 & 25.26 & 23.73 \\
 & ISTY \cite{DL_6} & -- & -- & -- & -- & 26.42 \\
 & ELFNet \cite{DL_9} & 29.18 & 22.54 & 28.16 & \textbf{31.63} & 27.87 \\
 & MANet \cite{DL_7} & \underline{29.51} & \underline{25.01} & \textbf{32.72} & 29.45 & 29.17 \\
 & Progressive Multi-Plane \cite{DL_10} & -- & -- & -- & -- & \underline{29.70} \\
 & LF-PyrNet \cite{DL_13} & -- & -- & -- & -- & 27.41 \\
\cmidrule{2-7}

 & Ours & 24.73 & 24.28 & 28.66 & 29.96 & 26.91 \\
\midrule
\multirow{15}{*}{{SSIM $\uparrow$}} 
 & All-In-Focus \cite{single_image_23} & 0.636 & 0.568 & 0.656 & 0.569 & 0.587 \\
 & RFR \cite{single_image_20} & 0.539 & 0.660 & 0.857 & 0.616 & 0.668 \\
 & Coordfill \cite{single_image_22} & 0.529 & 0.675 & 0.864 & 0.869 & 0.734 \\
 & GIP \cite{single_image_5} & 0.407 & 0.683 & 0.705 & 0.876 & 0.668 \\
 & PUT \cite{single_image_10} & 0.635 & 0.810 & 0.935 & 0.824 & 0.801 \\
 & DeOccNet \cite{DL_3} & 0.655 & 0.696 & 0.802 & 0.606 & 0.690 \\
 & Mask4D \cite{DL_4} & 0.822 & 0.817 & 0.855 & 0.789 & 0.804 \\
 & Zhang et al. \cite{DL_5} & 0.696 & 0.775 & 0.876 & 0.807 & 0.789 \\
 & ISTY \cite{DL_6} & -- & -- & -- & -- & 0.836 \\
 & ELFNet \cite{DL_9} & 0.851 & \underline{0.870} & \underline{0.914} & 0.779 & 0.854 \\
 & MANet \cite{DL_7} & 0.854 & \textbf{0.894} & \textbf{0.941} & 0.807 & 0.874 \\
 & Progressive Multi-Plane \cite{DL_10} & -- & -- & -- & -- & 0.872 \\
 & LF-PyrNet \cite{DL_13} & -- & -- & -- & -- & 0.873 \\
\cmidrule{2-7}

 & Ours & \textbf{0.865} & 0.854 & 0.910 & \textbf{0.904} & \textbf{0.883} \\
\bottomrule
\end{tabular*}
\end{table}

\subsection{Qualitative Results}

\begin{figure}
\centering
\includegraphics[width=\columnwidth]{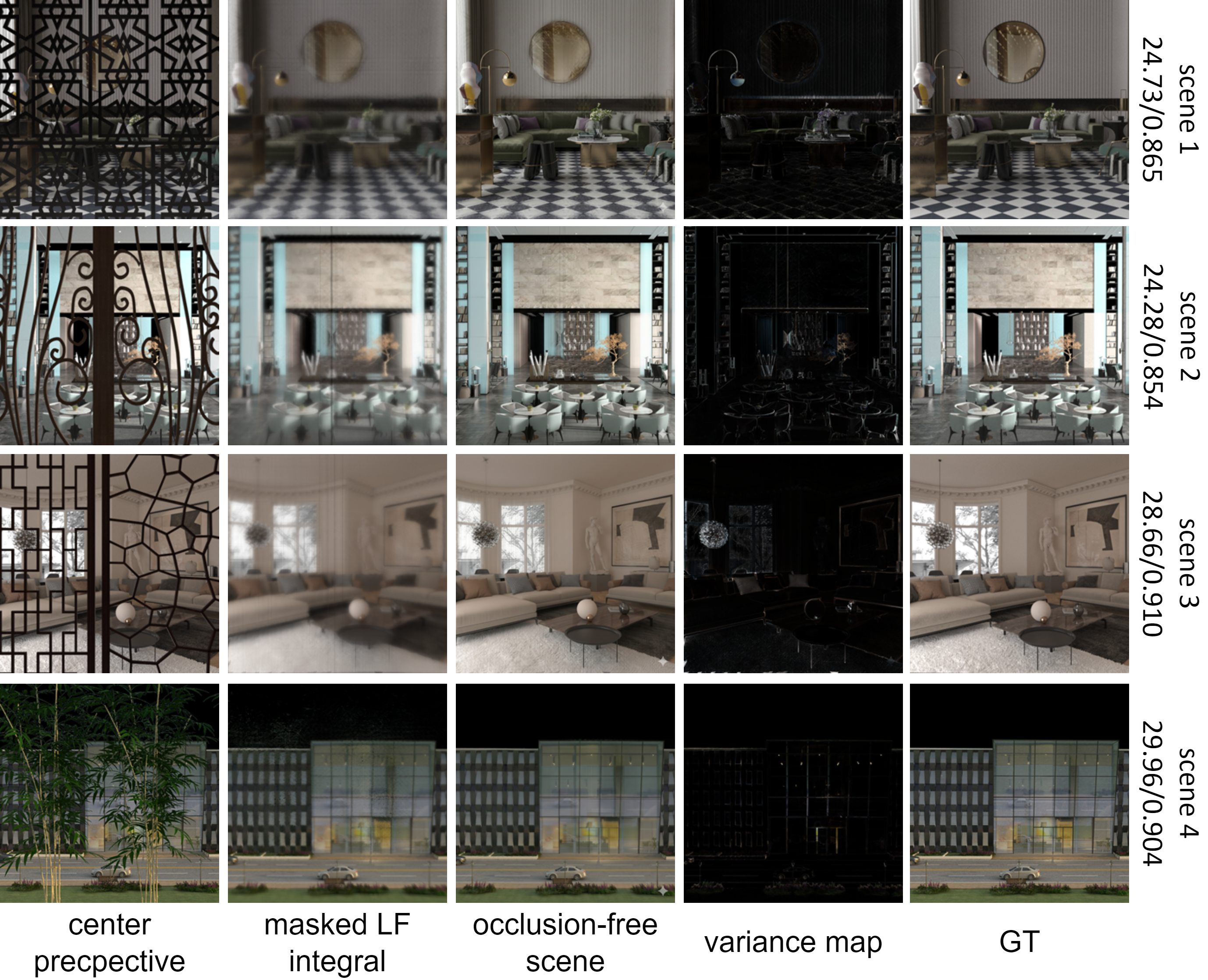}
\caption{Qualitative comparison of occlusion suppression and scene reconstruction results across 4-Syn benchmark scenes. From left to right: center perspective view, masked LF integral image, reconstructed occlusion-free scene using VLM, corresponding variance map of multiple VLM responses, and ground truth (GT). The center LF views are heavily degraded by foreground occluders, while the proposed framework suppresses occlusion effects and recovers the hidden scene structure. Variance maps highlight regions of reconstruction uncertainty, where lower variance indicates more consistent predictions. Quantitative performance metrics for each scene are shown on the right.}
\label{syn_4}
\end{figure}

Figure \ref{syn_4} presents a visualization of the 4-Syn benchmark scenes, showing a visual comparison between the LF integral image, ground truth, center perspective view, and the occlusion free image produced by our approach. The figure highlights the effectiveness of our method in reconstructing occlusion-free scenes with improved visual quality.

Additionally, the variance map illustrates the pixel-wise variation across multiple VLM responses, which reflects the degree of hallucination introduced by the model during the reconstruction process.

\begin{figure*}[!t]
\centering
\includegraphics[width=\textwidth]{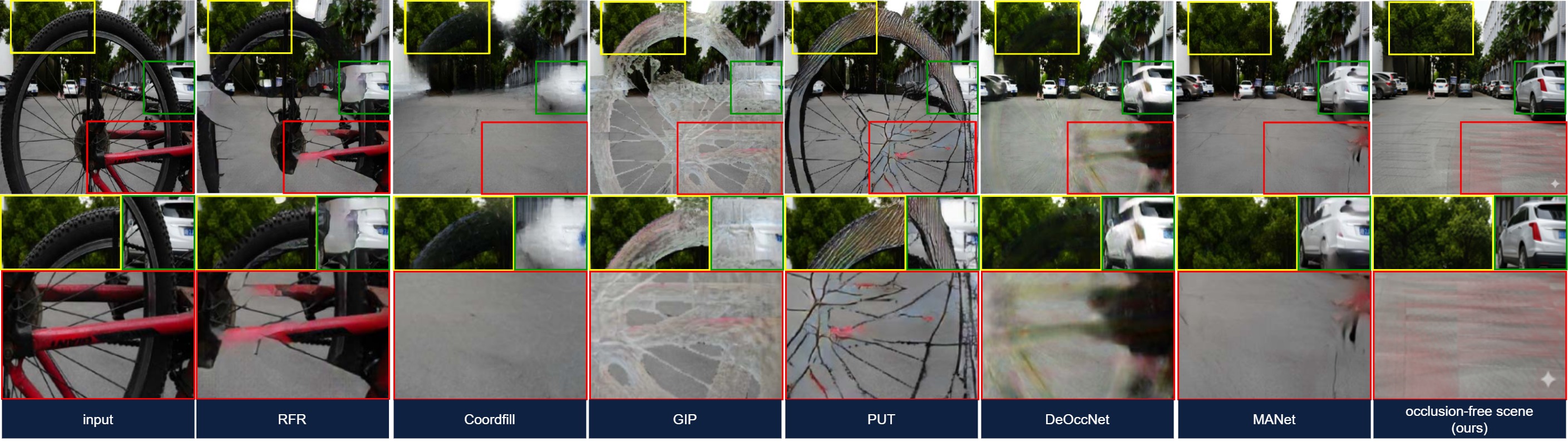}
\caption{Qualitative comparison on real-world data between the proposed framework and state-of-the-art methods. The proposed approach demonstrates strong generalization capability, particularly in the enlarged local regions, where it achieves improved preservation of structural details, reduced artifacts, and more coherent scene reconstruction.}
\label{real_1}
\end{figure*}

Figure~\ref{real_1} presents a qualitative comparison on real-world data between the proposed framework and state-of-the-art methods using the same prompt employed for the 4-Syn scenes. The results demonstrate the strong generalization capability of the proposed approach under real-world conditions. In particular, the enlarged local regions highlight its ability to preserve structural details, suppress reconstruction artifacts, and produce more coherent and visually plausible scene reconstructions compared with competing methods.

\begin{figure}
\centering
\includegraphics[width=\columnwidth]{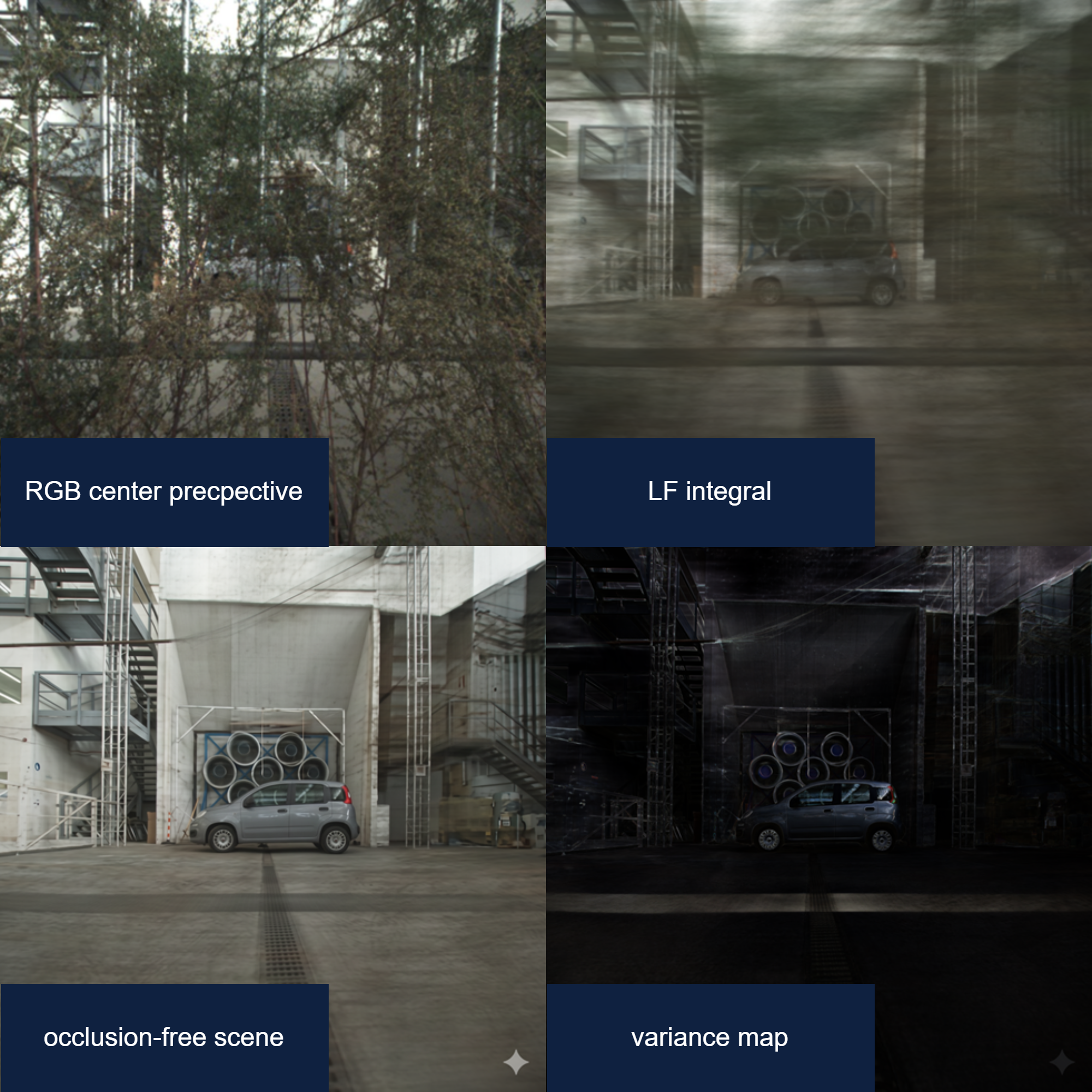}
\caption{Example of the proposed occlusion suppression and reconstruction process under dense vegetation occlusion. Top-left: RGB center perspective view, where the target scene is heavily obscured by foreground vegetation. Top-right: LF integral image with reduced foreground interference. Bottom-left: reconstructed occlusion-free  generated by the proposed framework, recovering underlying scene content and structural details. Bottom-right: corresponding variance map indicating reconstruction uncertainty, where brighter regions correspond to areas of greater prediction ambiguity.}
\label{robot}
\end{figure}

Fig.~\ref{robot} demonstrates a potential application of the proposed framework in robotic perception and scene understanding under challenging visibility conditions \cite{aos_3}. In natural and unstructured environments, robotic systems frequently encounter severe visual degradation from vegetation, clutter, or partial occlusions, reducing the reliability of downstream tasks such as navigation, obstacle avoidance, target identification, and path planning.

By suppressing foreground occlusions and recovering degraded scene information, the proposed approach enhances visibility of previously obscured regions, providing a more complete scene representation. This enables more informative visual observations and a more accurate understanding of environmental structure and object relationships, supporting more reliable scene interpretation and decision-making in robotic systems operating under limited visibility.

\begin{figure*}[!t]
\centering
\includegraphics[width=\textwidth]{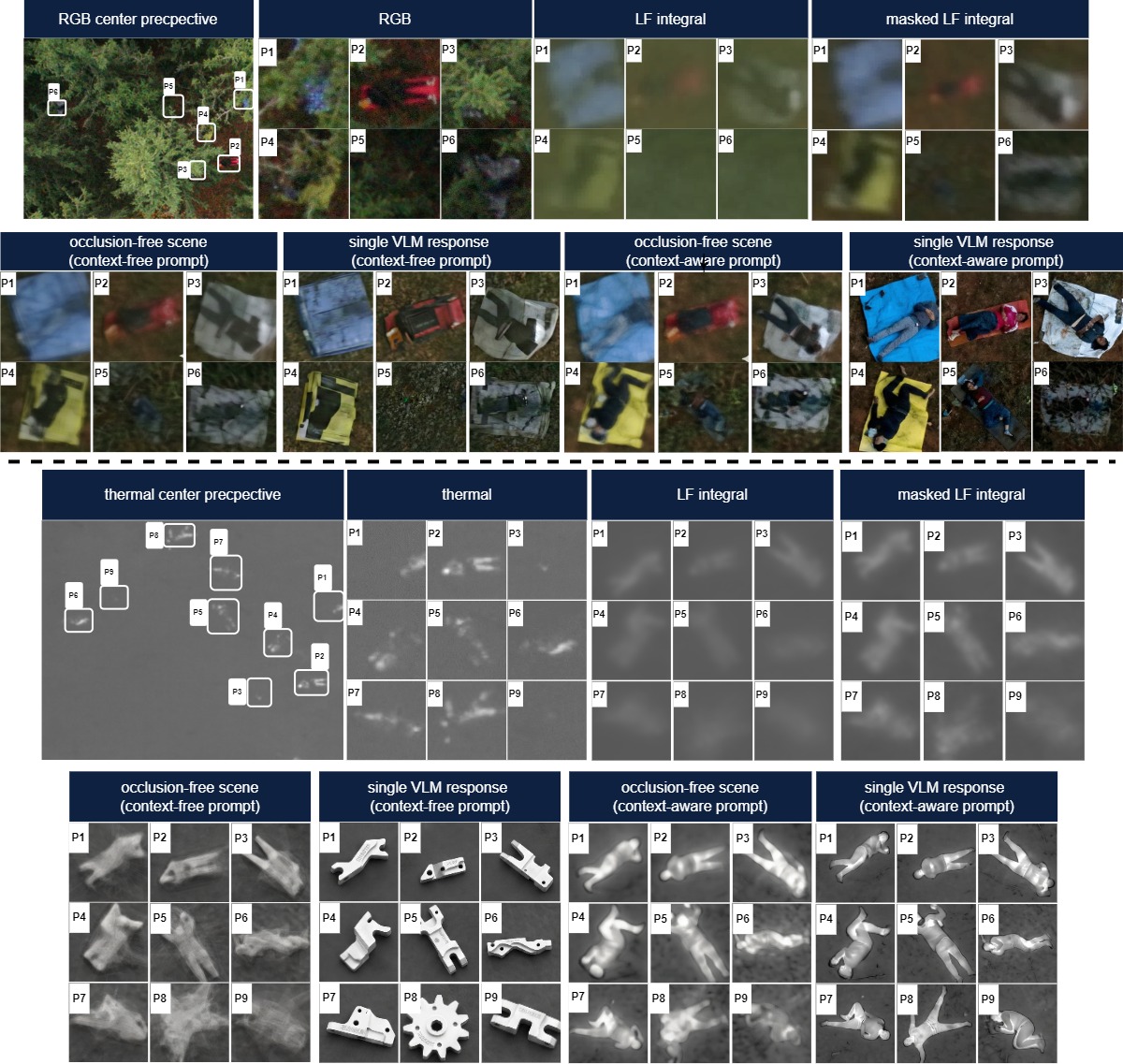}
\caption{Real-world search-and-rescue scenario under severe vegetation occlusion for RGB (top) and thermal (bottom) modalities, comparing center perspective view, LF integral image, masked LF integral image, occlusion-free reconstruction, and VLM responses under context-free and context-aware prompting. The proposed framework suppresses occlusion artifacts and enhances target visibility in both modalities. Close-ups (P1–P6 for RGB, P1–P9 for thermal) show context-aware prompting producing more coherent, semantically consistent reconstructions, improving recovery of obscured human targets while reducing hallucination compared to context-free prompting.}
\label{real_2}
\end{figure*}

Figure ~\ref{real_2} presents a real-world search-and-rescue scenario under severe vegetation occlusion for both RGB and thermal modalities, similar to challenging visibility conditions encountered in aerial rescue operations, comparing context-aware and context-free prompting strategies. The results demonstrate the strong generalization capability of the proposed framework in enhancing LFI reconstructions through effective suppression of occlusion artifacts and residual blur. Furthermore, incorporating contextual information substantially improves the recovery of obscured targets and produces more semantically consistent scene representations. This effect is particularly evident in the close-ups (P1–P6) for RGB and (P1–P9) for thermal modalities, where context-aware prompting yields improved structural coherence and more recognizable target characteristics.

For the context-aware setting, the VLM was provided with the following prompt:

\textit{``This is an aerial \(X\) image captured by a drone at approximately 35 m altitude during a search-and-rescue mission, showing people lying on a mat on the ground. Please remove the blur from the image. The processed image should remain aligned with the input image.''}

where \(X\)  denotes either RGB or thermal depending on the employed modality. Providing scene-specific contextual information enables the VLM to better infer the semantic structure of the environment and align the reconstruction process with the intended restoration objective. Consequently, the generated outputs exhibit improved reconstruction quality in both RGB and thermal modalities. Notably, the RGB occlusion-free scene reconstruction successfully recovers a person (see P6) who is not clearly distinguishable in either the original LF integral image or the masked LF integral image. Similarly, the reconstructed occlusion-free thermal image provides improved visualization of all persons compared to the thermal center view, LF integral image, and masked LF integral close-ups.

In contrast, when only a context-free prompt was employed,

\textit{``Please remove the blur from the image. The processed image should remain aligned with the input image.''}

the reconstructed outputs exhibited generative artifacts in both RGB and thermal modalities. Specifically, the VLM introduced objects such as books and plastic boxes that were inconsistent with the actual scene content, indicating hallucinated structures unsupported by the observed measurements.

These results suggest that incorporating scene-specific contextual information provides an effective mechanism for guiding VLM-based reconstruction toward semantically meaningful and physically consistent outputs while reducing hallucination artifacts in real-world LF occlusion removal scenarios.



\section{discussion}

The real-world results show that the proposed framework improves visibility of heavily occluded regions while preserving global scene structure. Integrating semantic vision reasoning enhances degraded regions and improves interpretability where conventional LF integral images remain blurry or incomplete. Since real-world scenarios lack occlusion-free ground truth, quantitative metrics such as PSNR and SSIM are infeasible, evaluation therefore focuses on qualitative structural consistency, visibility enhancement, and semantic plausibility.

The reconstructed outputs should not be interpreted as fully physically verified scene recovery. Because the VLM introduces learned semantic priors during refinement, portions of the generated content may be inferred rather than directly observed. The framework is therefore best viewed as a visibility enhancement and scene interpretation approach rather than a guaranteed physically accurate reconstruction system.
Although VLM refinement improves perceptual quality and structural coherence, its generative nature can introduce hallucinated content, particularly in severely degraded regions with limited observations. Fig.~\ref{real_2} illustrates this: context-free prompting generates artificial objects inconsistent with the captured scene, underscoring the need for physically grounded constraints and contextual information during refinement.

To mitigate hallucination, the framework uses two complementary strategies: the VLM refines the LF integral image rather than generating from scratch, improving measurement consistency, and multiple independently generated samples are averaged with the all-in-focus integral image to suppress inconsistent hallucinations while preserving stable structures. Hallucination-free reconstruction cannot be fully guaranteed, particularly under extremely limited visibility, so the framework should be applied cautiously in safety-critical applications.
Computational performance depends mainly on the occlusion masking approach and the selected VLM. The LFI stage runs near real-time on an RTX 5090 (several milliseconds). Semantic refinement is the primary bottleneck: Gemini 3.1 took approximately 23 seconds per image (varying with server load), while Qwen-Image took approximately 9 minutes locally on an RTX 5090. Averaging multiple samples further increases inference time proportionally, and masking cost varies with the segmentation or depth estimation method used, from real-time lightweight approaches to slower learning-based ones.

Future work includes integrating event cameras to improve robustness under challenging illumination and visibility, and improving semantic refinement efficiency to enable real-time deployment on autonomous aerial robotic systems in complex environments.

\section{conclusion}
In this article, we proposed a vision-reasoning-enhanced light field occlusion removal framework for robust scene recovery, combining the visibility enhancement capabilities of LFI with the semantic reasoning abilities of modern VLMs to recover scene information under severe occlusion. By integrating multi-view observations with learned semantic priors, the framework enables improved reconstruction of degraded and partially observed content.

Experimental results on synthetic and real-world datasets show that our method consistently outperforms state-of-the-art approaches, both quantitatively and qualitatively, while exhibiting strong generalization to unstructured acquisition settings without requiring strict viewpoint configurations. We also observe that context-aware prompting reduces hallucination and improves VLM output quality compared to context-free prompting. These results highlight the approach's applicability across domains such as search-and-rescue and exploratory robotic navigation, where reliable perception under challenging visibility conditions is essential.

 

\newpage
\bibliographystyle{IEEEtran}
\bibliography{sn-bibliography}

\end{document}